\documentclass[conference]{IEEEtran}
\IEEEoverridecommandlockouts
\usepackage{cite}
\usepackage{amsmath,amssymb,amsfonts}
\usepackage{algorithmic}
\usepackage{graphicx}
\usepackage{textcomp}
\usepackage{xcolor}
\def\BibTeX{{\rm B\kern-.05em{\sc i\kern-.025em b}\kern-.08em
    T\kern-.1667em\lower.7ex\hbox{E}\kern-.125emX}}
    
\usepackage{mwe}
\usepackage[markcase=noupper
]{scrlayer-scrpage}
\ohead{}
\begin{document}
\IEEEoverridecommandlockouts
\IEEEpubid{\begin{minipage}[t]{\textwidth}\ \\[10pt]
        \centering\normalsize{U.S. Government work not protected by U.S. copyright}
\end{minipage}} 

\title{An Approach to Partial Observability in Games: Learning to Both Act and Observe\\
\thanks{Distribution Statement A: Approved for public release. Distribution is unlimited.}
}

\author{\IEEEauthorblockN{1\textsuperscript{st} Elizabeth Gilmour}
\IEEEauthorblockA{\textit{Naval Center for Applied Research in A.I.} \\
\textit{U.S. Naval Research Laboratory}\\
Washington D.C., U.S.A. \\
elizabeth.gilmour@nrl.navy.mil}
\and
\IEEEauthorblockN{2\textsuperscript{nd} Noah Plotkin}
\IEEEauthorblockA{\textit{Oberlin College} \\
Oberlin, OH U.S.A. \\
nplotkin@oberlin.edu}
\and
\IEEEauthorblockN{3\textsuperscript{rd} Leslie N. Smith}
\IEEEauthorblockA{\textit{Naval Center for Applied Research in A.I.} \\
\textit{U.S. Naval Research Laboratory}\\
Washington D.C., U.S.A. \\
leslie.smith@nrl.navy.mil}

}

\maketitle

\begin{abstract}
Reinforcement learning (RL) is successful at learning to play games where the entire environment is visible. However, RL approaches are challenged in complex games like Starcraft II and in real-world environments where the entire environment is not visible. In these more complex games with more limited visual information, agents must choose where to look and how to optimally use their limited visual information in order to succeed at the game. We verify that with a relatively simple model the agent can learn where to look in scenarios with a limited visual bandwidth. We develop a method for masking part of the environment in Atari games to force the RL agent to learn both where to look and how to play the game in order to study where the RL agent learns to look. In addition, we develop a neural network architecture and method for allowing the agent to choose where to look and what action to take in the Pong game. Further, we analyze the strategies the agent learns to better understand how the RL agent learns to play the game.
\end{abstract}


\section{Introduction}
In recent years, reinforcement learning (RL) agents have made significant progress learning to play games \cite{videogames}. These RL agents are able to directly process raw pixels from the game environment as input, and from this input, they learn strategies for game play. The agents are able to learn both the perception tasks and the decision tasks needed for successful game play. Learning to play games from visual information is common across many video games, from the simple Atari games \cite{atari} to the complex Starcraft II \cite{b1,b2,AlphaStar,SC2}.

In some games, such as the Atari 2600 games, the agent can view the entire environment, and so the agent has perfect information about all aspects of the game. RL agents have shown excellent performance at these games \cite{atari,b1}. In cases in which agents cannot see the entire game at one time and do not have perfect information, strategies are needed to deal with the challenge of learning where to look and how to make decisions with limited visual information. In developing an agent to play Starcraft II, for example, researchers used supervised training and imitation learning to teach the agent successful strategies \cite{AlphaStar}.

Making intelligent choices about what visual information to acquire is not trivial and has applications beyond games, such as in robotic navigation. It cannot be assumed that an RL agent will know where to look and which visual imagery to acquire \cite{b3},\cite{b4}. Learning where to look, how to acquire useful visual information, and how to make use of available visual information are three closely connected and challenging problems in RL.

Previous work with RL agents has demonstrated the ability to play video games with limited observations \cite{neural evolution,PO21} by learning representations containing task relevant information \cite{representation}.  However, these methods primarily relied on attention or other complex network architectures, such as recurrent neural networks \cite{recurrent}, query and keys \cite{neural evolution}, positional encoding, and transformers \cite{transformers}. In contrast to these complicated RL models, a simple model allows for easier understanding and analysis of how RL agents make decisions regarding where to look in order to play video games.  RL models are not easily interpretable for humans \cite{b6} so a simple model can increase understanding of how a model learns to accomplish this task. 

In this paper we introduce partial observability into a game to provide a testbed for understanding how agents learn which visual information to acquire. We show that a relatively simple convolutional neural network (CNN) is able to learn where to look. This work provides new understanding about what information the agent needs to make decisions. 

\section{Related Works}

While very effective, RL models are not easily interpreted to make sense of what information is used and why actions are taken \cite{b6}. Understanding what visual information is important for playing a game is commonly studied through attention and saliency.

Attention models allow the agent to attend to a small area of the image \cite{b6}. These models often use bottlenecks or artificial retinas to limit the amount of visual information the model can use to make decisions \cite{neural evolution},  \cite{recurrent}, \cite{b6}. Observing which areas the models attend to improves interpretability \cite{neural evolution}.  The attention agents find the important information and ignore unnecessary visual information, and as a result, they can succeed in modified environments and different contexts \cite{neural evolution}.  Unlike previous work with attention modules, we show that a simple RL agent can choose where to look, and given that glimpse of the observation space, it can learn to reach a high score in a video game.

Saliency models increase interpretability by creating heat-map like outputs that allow for visualization of the importance of different parts of the image \cite{b5}, \cite{b7}. These visualizations can be used to understand the importance of specific parts of the image for policy decisions \cite{b5} as well as understanding the agent’s internal representation and behavior \cite{b7}. 

For both saliency and attention models, the RL agent usually sees the whole environment. In this paper, the agent can only see a fraction of the environment and must choose what to view to gain the visual information needed to succeed at the game. While this approach to partial observability in RL has certain similarities to saliency and attention, this work differs in that the architecture is much simpler than agents with attention or self-attention mechanisms.

Perhaps the closest to our work is that of reconnaissance blind chess \cite{b8} where the agent must choose what fraction of the chess board to view in order to win the game.  In this paper, a similar concept is adapted to learning where to look when playing video games.

\section{Approach}

To test the performance of RL models in a simple game with limited visual information, we implement partial observability in the Atari game Pong. To do so, we artificially reduce the area that the model can see during each time step. We then develop a neural network that can handle the perception tasks as well as decision making for the partially observable game. Further, we explore how the model learns to play the game and chooses where to look.

To implement partial observability, we develop masks that occlude two-thirds of the environment, allowing the agent to view only one-third of the screen at a time. The agent must choose not only what action to take in the game, but also which occlusion map to view. The occlusion maps force the model to not only make decisions based on limited visual information, but also create the challenge of learning which visual information to acquire to succeed at the game. The reward is still based only on the score in the game, not based on the model’s success at any specific strategy of acquiring visual information. 

Creating a simple game with limited visual information allows us to observe what strategies the model learns. It also allows for comparison with state-of-the-art fully observable games. In contrast to the approach in \cite{b2}, we do not attempt to teach the RL agent any particular strategies for choosing where to look but instead study the strategies the agent learns on its own.

\subsection{Deep Q Network}

The deep Q network (DQN) is a reinforcement learning algorithm that has shown great success at simple video games such, such as the Atari games \cite{b1}. The DQN works for games with a discrete action space. 

The DQN is able to learn from raw pixels using a CNN. As the DQN interacts with an environment, such as an Atari game, the model chooses an action at each time step with the goal of maximizing the score in the game.

\subsection{Atari Pong}

Atari Pong is one of the Atari 2600 games. The agent plays a modified ping pong game against the computer. The goal is to score points against the opponent. The lowest possible score in the game is -21 and the maximum possible score is 21, and the state-of-the-art score is 21. 

At each time step, the agent can choose to move the paddle up or down, or it can choose to take no action. The simple and discrete action space makes Pong a good testbed for new approaches.

\subsection{Occlusion}

\begin{figure}[htbp]
\begin{center}
\includegraphics[width=0.8\columnwidth]{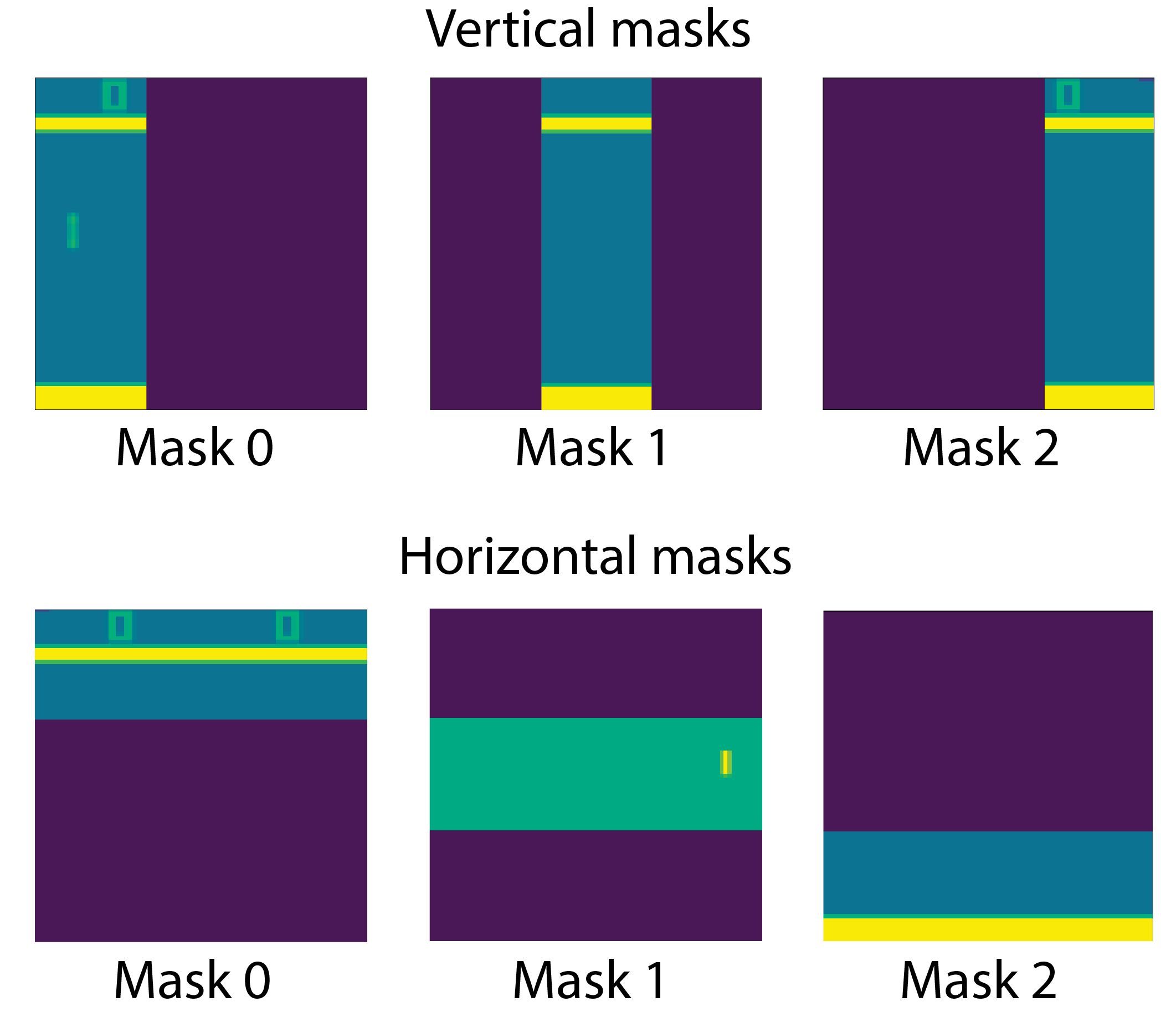}
\caption{The masks allow the model to only view one-third of the screen at any time. Both vertical and horizontal masks are used.}
\label{masks}
\end{center}
\end{figure}

Partial observability is implemented by limiting which part of the game the agent can view. As the agent learns from raw pixels, it is possible to apply masks to the input to the agent to obscure some of the pixels. We allowed the model to view one-third of the game at a time. We tested two different types of occlusions masks: horizontal where the model can choose to view the top third, middle third, or bottom third of the screen or vertical where the model can choose to view the right third, middle third, or left third of the screen (Fig.~\ref{masks}).
\begin{figure}[htbp]
\begin{center}
\includegraphics[width=0.9\columnwidth]{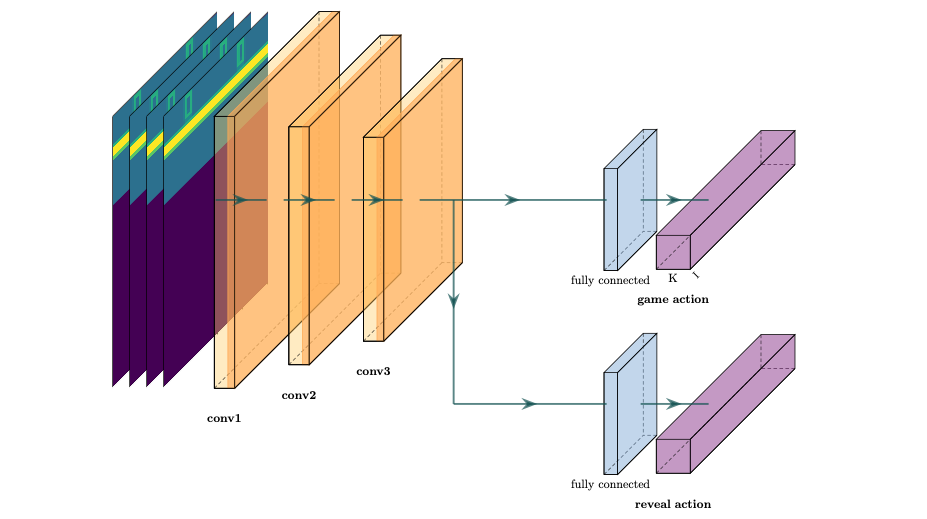}
\caption{The model architecture is based on a backbone CNN. The model then has two heads, one that makes the decision about the action to take in the game and a second that chooses where to look.}
\label{architecture}
\end{center}
\end{figure}
\subsection{Model Architecture}

The model used for the DQN is a CNN. The CNN can be trained to accomplish both perception and decision task. Four frames from the game are used as input to the model for each time step. To play partially-observable Atari Pong, it was necessary to create a new CNN that outputs two different types of actions—a choice of where to look and a choice of what action to take in the game. 
Applying RL models to larger and higher dimension action spaces is a challenging problem. Combining different types of actions leads to a combinatorial action space, the size of which rapidly increases when more actions are available. A solution to a combinatorial action space is the action branching architecture in which a shared representation feeds into several networks, each of which control one degree of freedom \cite{b8}.

Based on the action branching architecture, we developed a neural network architecture (Fig.~\ref{architecture}) that has a shared backbone of convolutional layers and two heads made up of more convolutional layers. Each head has a separate output. As a result, it was necessary to convert the two separate outputs into a single combinatorial action. The new action space is a discrete action space with every combination of action in the game and choice of occlusion mask. 

\section{Experiments and Results}

\begin{figure}[htbp]
\includegraphics[width=\columnwidth]{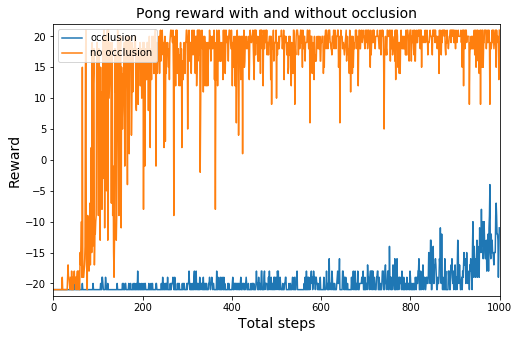}
\caption{The training reward for the fully observable game (shown in orange) rapidly increases with training. In the partially observable game (shown in blue) the training reward improves more slowly.}
\label{training}
\end{figure}

The model is tested with both the horizontal and vertical masks. To test its performance at the partially observable tasks, the model is trained for 1000 episodes. The training procedure is the same as for the standard DQN, but the model must be trained for longer because of the increased complexity of the larger action space. Little improvement in the score could be seen after 1000 episodes (Fig.~\ref{training}).

To improve performance and reduce the training time, we implemented curriculum learning. In curriculum learning, a model is first trained on an easier task, and once the model succeeds at the easier task, the more difficult task is introduced. To implement curriculum learning for partially observable Atari Pong, we begin by training the model on the fully observable game. Doing so allows the model to learn to play the game and score points. Only once the model is able to reach the maximum score of 21 are the occlusion masks added.

\begin{figure}[htbp]
\includegraphics[width=\columnwidth]{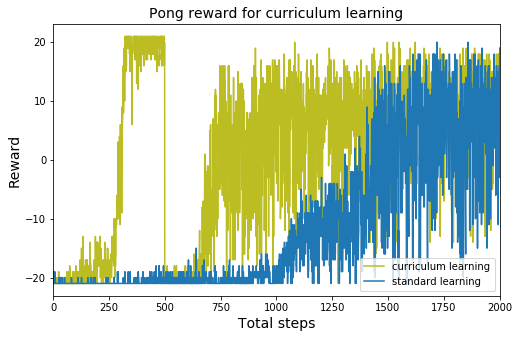}
\caption{The model using curriculum learning (shown in green) quickly learns to play the fully observable game. When occlusion is introduced after 500 steps, the model quickly adapts to the partially observable game.}
\label{curriculum}
\end{figure}

Fig.~\ref{curriculum} shows the improvement from the curriculum learning compared to the standard training. Either with or without curriculum learning, the model achieves a state-of-the-art score. The results show that this novel network architecture allows an RL agent to learn where to look in order to succeed at a task.

Analyzing where the model chooses to look can provide insight about the strategies that the model learns. Counting how often the model chooses each mask provides information about which areas on the screen are most important for succeeding at game play. Similar to saliency, this information shows what the model attends to as it plays the game.

When playing Atari with horizontal masks, the model chooses to look at each section with similar frequency, but in the situation with vertical masks, the model chooses to look at the right side the majority of the time (Table 1). This result suggests that the model needs to look at the location of its own paddle much of the time, either to see where its own paddle is or to defend against points scored by the opponent.

\begin{figure*}[ht]
\centering
\begin{center}
\includegraphics[width=2\columnwidth]{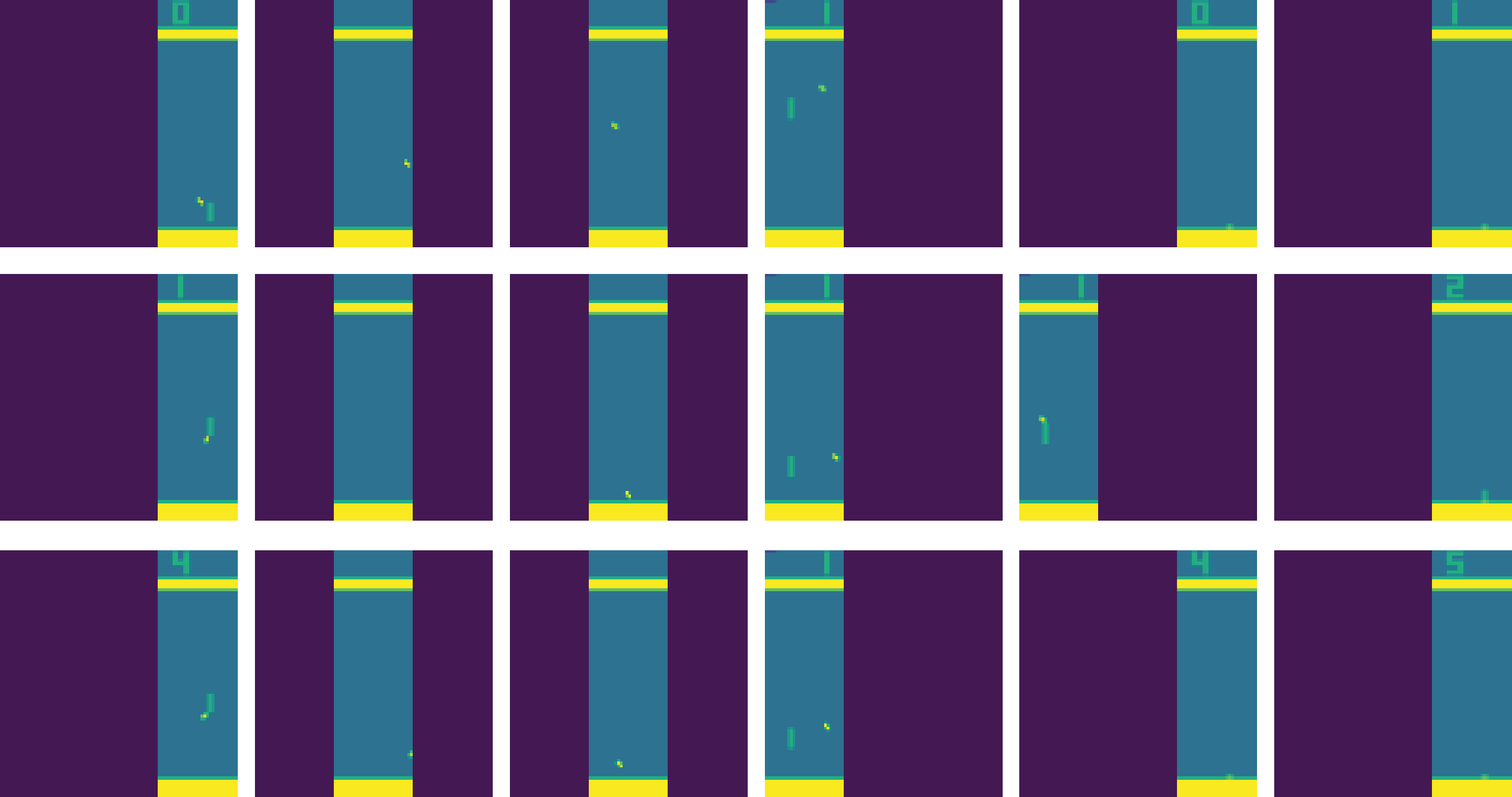}
\caption{Renderings of the masked Pong screen show the sequence of actions the model takes before scoring. After hitting the ball back to the opponent, the model chooses where to look in order to view the ball as it moves across the screen. After scoring, the model once again chooses to view its own paddle.}
\label{sequence}
\end{center}
\end{figure*}

\begin{table}[]
\caption{Mask choices during gameplay}
\begin{center}
\begin{tabular}{|l|l|l|l|}
\hline
\multicolumn{2}{|c|}{\textbf{Horizontal masks}} & \multicolumn{2}{c|}{\textbf{Vertical masks}} \\ \hline
\textbf{Mask}          & \textbf{Count}         & \textbf{Mask}        & \textbf{Count}        \\ \hline
Top                    & 467                    & Left                 & 525                   \\ \hline
Middle                 & 599                    & Middle               & 370                   \\ \hline
Bottom                 & 743                    & Right                & 1035                  \\ \hline
\end{tabular}
\label{tab1}

\end{center}
\end{table}

Viewing the model's game play provides additional information about the strategies learned. Fig.~\ref{sequence} shows renderings of the screen with vertical masks. The figure shows every fourth frame in the short period of time before the agent scores. In each sequence of frames, the model observes the ball as it moves across the screen, similar to a human’s strategy of “keeping your eye on the ball”.

\section{Discussion}

Our simple RL model successfully learns strategies for where to look in order to accomplish a task. The model does not require any predetermined strategy or supervised training but instead it is able to develop strategies in the course of learning to maximize the task reward function. Despite viewing only a portion of the screen, there is little reduction in performance compared to the fully observable game. The model can learn which visual information is most important to succeed at a game.

This work serves as a proof-of-concept for the challenge of training a model to choose both where to look and what action to take using the context in the the image that can be viewed. This work can be extended for more complicated video games including real time strategy games and for real-world scenarios like robotic navigation or acquisition of new imagery. Our work so far has only focused on testing the model on Atari Pong, but testing on more challenging games is left for future work.  We expect that the concepts uncovered here will carry over into these more complex domains.

\vspace{12pt}


\begin{thebibliography}{00}
\bibitem{videogames} Shao, Kun, Zhentao Tang, Yuanheng Zhu, Nannan Li, and Dongbin Zhao. "A survey of deep reinforcement learning in video games." arXiv preprint arXiv:1912.10944 (2019).
\bibitem{atari} Badia, Adrià Puigdomènech, Bilal Piot, Steven Kapturowski, Pablo Sprechmann, Alex Vitvitskyi, Zhaohan Daniel Guo, and Charles Blundell. "Agent57: Outperforming the atari human benchmark." In International Conference on Machine Learning, pp. 507-517. PMLR, 2020.
\bibitem{b1} V. Mnih, K. Kavukcuoglu, D. Silver, A. Graves, I. Antonoglou, D. Wierstra, and M. Riedmiller, “Playing Atari with deep reinforcement learning,” arXiv preprint arXiv:1312.5602, December 19 2013.
\bibitem{b2} V. Mnih, K. Kavukcuoglu, D. Silver, A.A. Rusu, J. Veness, M.G. Bellemare, A. Graves, M. Riedmiller, A.K. Fidjeland, G. Ostrovski, and S. Petersen, “Human-level control through deep reinforcement learning,” Nature, 518(7540):529-33, February 2015.
\bibitem{AlphaStar} Vinyals, Oriol, Igor Babuschkin, Wojciech M. Czarnecki, Michaël Mathieu, Andrew Dudzik, Junyoung Chung, David H. Choi et al. "Grandmaster level in StarCraft II using multi-agent reinforcement learning." Nature 575, no. 7782 (2019): 350-354.
\bibitem{SC2} Vinyals, Oriol, Timo Ewalds, Sergey Bartunov, Petko Georgiev, Alexander Sasha Vezhnevets, Michelle Yeo, Alireza Makhzani et al. "Starcraft ii: A new challenge for reinforcement learning." arXiv preprint arXiv:1708.04782 (2017).
\bibitem{b3} D. Jayaraman and K. Grauman, “Learning to look around: Intelligently exploring unseen environments for unknown tasks,” In Proceedings of the IEEE Conference on Computer Vision and Pattern Recognition, (pp. 1238-1247) 2018.
\bibitem{b4} S. Mousavi, M. Schukat, E. Howley, A. Borji, and N. Mozayani, “Learning to predict where to look in interactive environments using deep recurrent Q-learning,” arXiv preprint arXiv:1612.05753, December 2016.
\bibitem{neural evolution}Y.  Tang, D. Nguyen, and D. Ha, "Neuroevolution of self-interpretable agents," In Proceedings of the 2020 Genetic and Evolutionary Computation Conference, (pp. 414-424), June 25 2020.
\bibitem{PO21}  Ramicic, Mirza, and Andrea Bonarini. "Uncertainty Maximization in Partially Observable Domains: A Cognitive Perspective." arXiv preprint arXiv:2102.11232 (2021).
\bibitem{representation} Zhang, Amy, Rowan McAllister, Roberto Calandra, Yarin Gal, and Sergey Levine. "Learning invariant representations for reinforcement learning without reconstruction." arXiv preprint arXiv:2006.10742 (2020).
\bibitem{recurrent} V. Mnih, N. Heess, and A. Graves. "Recurrent models of visual attention." In Advances in neural information processing systems, pp. 2204-2212. 2014.
\bibitem{transformers} Khan, Salman, Muzammal Naseer, Munawar Hayat, Syed Waqas Zamir, Fahad Shahbaz Khan, and Mubarak Shah. "Transformers in vision: A survey." arXiv preprint arXiv:2101.01169 (2021).
\bibitem{b5} D. Nikulin, A. Ianina, V. Aliev, and S. Nikolenko, “Free-lunch saliency via attention in Atari agents,” In 2019 IEEE/CVF International Conference on Computer Vision Workshop (ICCVW) (pp. 4240-4249), October 2019.
\bibitem{b6} A. Mott A, D. Zoran, M. Chrzanowski, D. Wierstra, and D.J. Rezende DJ, “Towards interpretable reinforcement learning using attention augmented agents,” arXiv preprint arXiv:1906.02500, June 2019.
\bibitem{b7} A. Atrey, K. Clary, and D. Jensen. “Exploratory not explanatory: Counterfactual analysis of saliency maps for deep reinforcement learning,” arXiv preprint arXiv:1912.05743, December 2019.
\bibitem{b8} A. Tavakoli, F. Pardo, and P. Kormushev, “Action branching architectures for deep reinforcement learning,” In Proceedings of the AAAI Conference on Artificial Intelligence, (Vol. 32, No. 1), April 2018.
\bibitem{b9} S. Narvekar, B. Peng, M. Leonetti, J. Sinapov, M.E. Taylor, and P. Stone, “Curriculum learning for reinforcement learning domains: A framework and survey,” Journal of Machine Learning Research, 1;21(181):1-50, January 2020.
\end{thebibliography}
\end{document}